\documentclass[10pt,twocolumn,letterpaper]{article}

\usepackage{cvpr}              

\usepackage{graphicx}
\usepackage{amsmath}
\usepackage{amssymb}
\usepackage{booktabs}
\usepackage{xcolor}

\usepackage[pagebackref,breaklinks,colorlinks]{hyperref}

\usepackage[capitalize]{cleveref}
\crefname{section}{Sec.}{Secs.}
\Crefname{section}{Section}{Sections}
\Crefname{table}{Table}{Tables}
\crefname{table}{Tab.}{Tabs.}

\definecolor{color1}{RGB}{179,226,205}
\definecolor{color2}{RGB}{253,205,172}
\definecolor{color3}{RGB}{203,213,232}
\definecolor{color4}{RGB}{244,202,228}
\definecolor{color5}{RGB}{230,245,201}
\definecolor{color6}{RGB}{255,242,174}

\definecolor{fontcolor1}{RGB}{228,26,28}
\definecolor{fontcolor2}{RGB}{55,126,184}
\definecolor{fontcolor3}{RGB}{77,175,74}
\definecolor{fontcolor4}{RGB}{152,78,163}
\definecolor{fontcolor5}{RGB}{255,127,0}

\def\cvprPaperID{560} 
\def\confName{CVPR}
\def\confYear{2023}

\usepackage{cuted}
\usepackage{capt-of}

\newif\ifnotes
\notestrue 
\ifnotes
\newcommand{\vjnote}[1]{\textcolor{magenta}{\textbf{Varun}: #1 }}

\newcommand{\bdnote}[1]{\textcolor{red}{Brendan: #1}}
\newcommand{\arnote}[1]{\textcolor{orange}{Arjun: #1}}
\else
\newcommand{\vjnote}[1]{}
\newcommand{\bdnote}[1]{}
\newcommand{\arnote}[1]{}
\fi

\begin{document}

\title{MetaCLUE: Towards Comprehensive Visual Metaphors Research}

\author{Arjun R. Akula$^*$, Brendan Driscoll$^*$, Pradyumna Narayana, Soravit Changpinyo,\\
Zhiwei Jia, Suyash Damle, Garima Pruthi, Sugato Basu, \\
Leonidas Guibas, William T. Freeman, Yuanzhen Li, Varun Jampani$^*$  \\
Google
}

\twocolumn[{%
\renewcommand\twocolumn[1][]{#1}%
\maketitle
\begin{center}
\centering
\includegraphics[width=\textwidth]{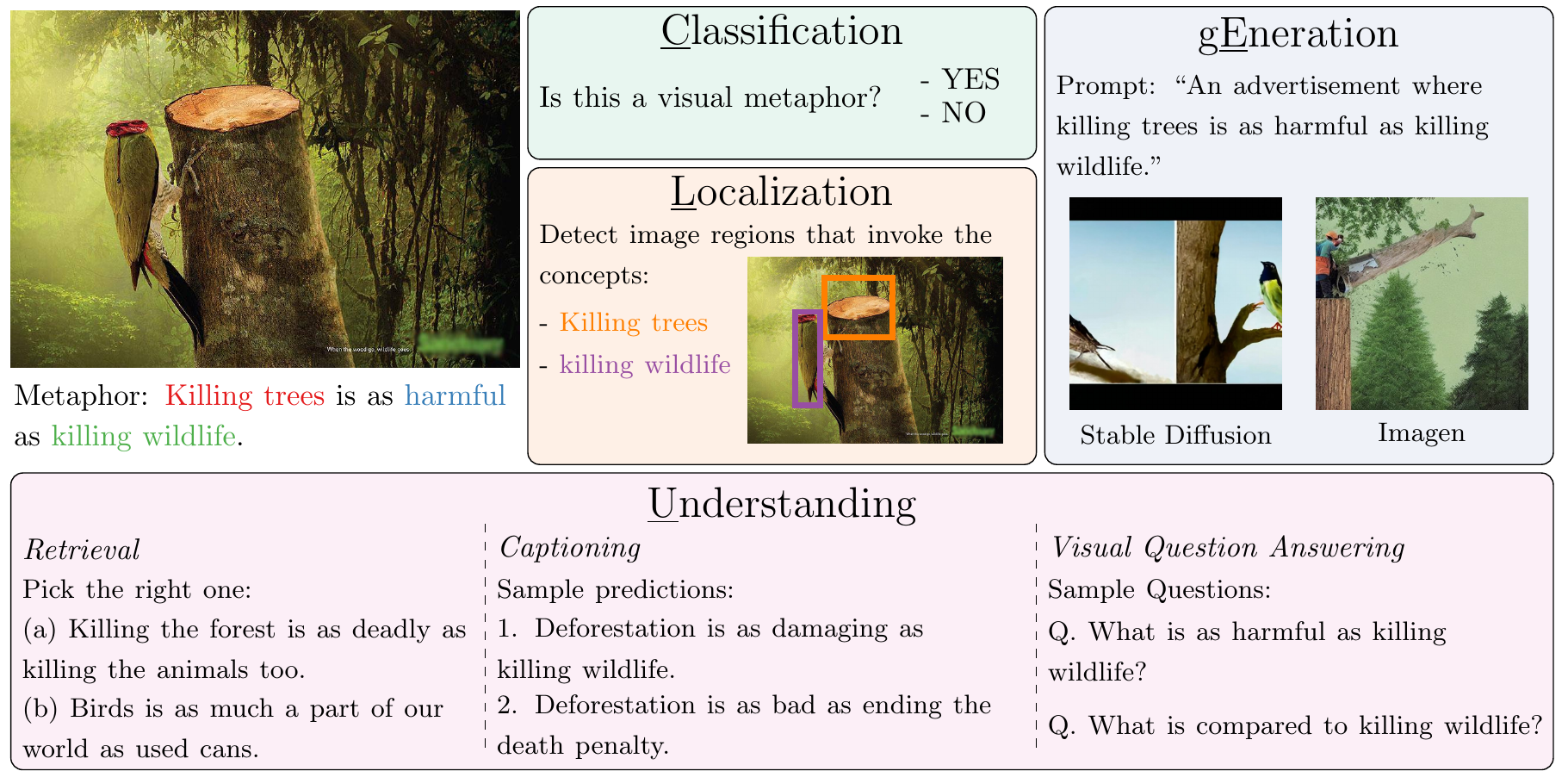}
\captionof{figure}{{\small With \textbf{MetaCLUE}, we introduce several interesting tasks related to visual metaphors. We collect metaphor annotations (objects, abstract concepts, relationships and object boxes) for evaluating existing models on these tasks. Specifically we perform a comprehensive evaluation of vision and language models on four different tasks (\textbf{C}lassification, \textbf{L}ocalization, \textbf{U}nderstanding, and g\textbf{E}neration). Comprehensive experiments in this work show that state-of-the-art techniques mostly focus on literal interpretation and perform poorly in understanding and generation of metaphor images.}\vspace{-1mm}
\label{fig:teaser}}
\end{center}%
}]
\def\thefootnote{*}\footnotetext{Equal Contribution}\def\thefootnote{\arabic{footnote}}
\begin{abstract}
\vspace{-3mm}
Creativity is an indispensable part of human cognition and also an inherent part of how we make sense of the world. Metaphorical abstraction is fundamental in communicating creative ideas through nuanced relationships between abstract concepts such as feelings. While computer vision benchmarks and approaches predominantly focus on understanding and generating literal interpretations of images, metaphorical comprehension of images remains relatively unexplored. Towards this goal, we introduce MetaCLUE, a set of vision tasks on visual metaphor.
We also collect high-quality and rich metaphor annotations (abstract objects, concepts, relationships along with their corresponding object boxes) as there do not exist any datasets that facilitate the evaluation of these tasks. We perform a comprehensive analysis of state-of-the-art models in vision and language based on our annotations, highlighting strengths and weaknesses of current approaches in visual metaphor \underline{c}lassification, \underline{l}ocalization, \underline{u}nderstanding (retrieval, question answering, captioning) and g\underline{e}neration (text-to-image synthesis) tasks. We hope this work provides a concrete step towards developing AI systems with human-like creative capabilities. Project page: \url{https://metaclue.github.io}


\end{abstract}
\vspace{-5mm}

\section{Introduction}
\label{sec:intro}
\begin{quote}
    \textit{``Metaphor is pervasive in everyday life ... Our ordinary conceptual system, in terms of which we both think and act, is fundamentally metaphorical in nature.''} --- Lakoff \& Johnson \cite{lakoff2008metaphors}
\end{quote}

Creativity is a process of generating a new perspective on a problem or a situation. Metaphorical thinking has been recognized as a key and powerful mechanism of creativity ~\cite{lakoff1993contemporary,lubart1997emotion,veale2016metaphor}. Humans engage metaphors in their creative thinking process as strategies to link or blend concepts, or to view a concept from a target domain in terms of another, apparently dissimilar concept from a source domain~\cite{lakoff2008metaphors}. Metaphors also provide a sophisticated tool for nuanced human communication. Let us take a closer look at the structure of metaphors -- and especially visual metaphors.

\vspace{0.1mm}
\noindent \textbf{Metaphors}\footnote{Grammarians distinguish a metaphor ``A is B'' from a simile ``A is like B''. In our work we use ``metaphor" to encompass both variants.}
are a cognitive construct in which a concept is compared to a seemingly unrelated concept via some shared attribute. Take as an example `This car is a cheetah', where `This car' is compared to `a cheetah' in terms of speed. Metaphors have a simple syntactic structure of `A is B' where A is referred to as the \textbf{primary concept} and B as the \textbf{secondary concept}. The implied analogy in a metaphor is of the form: `(primary concept) is as (relationship)\footnote{We use the word `relationship' to denote the shared property of primary and secondary concepts, usually adjectives or adjectival phrases.} as (secondary concept)' and often involves an attribute transfer from the secondary to the primary concept. Some examples include `This phone is as fast as a rocket', `Cigarettes are as harmful as bullets' etc. The primary and secondary concepts are usually unrelated at a glance, resulting in an element of surprise and creativity in metaphorical expressions. Despite following such simple structure, metaphors are quite powerful in conveying creative ideas.
Metaphors are pervasive in all forms of communication, such as speech, text, visual etc.

\begin{figure*}
\centering
\includegraphics[width=0.97\linewidth]{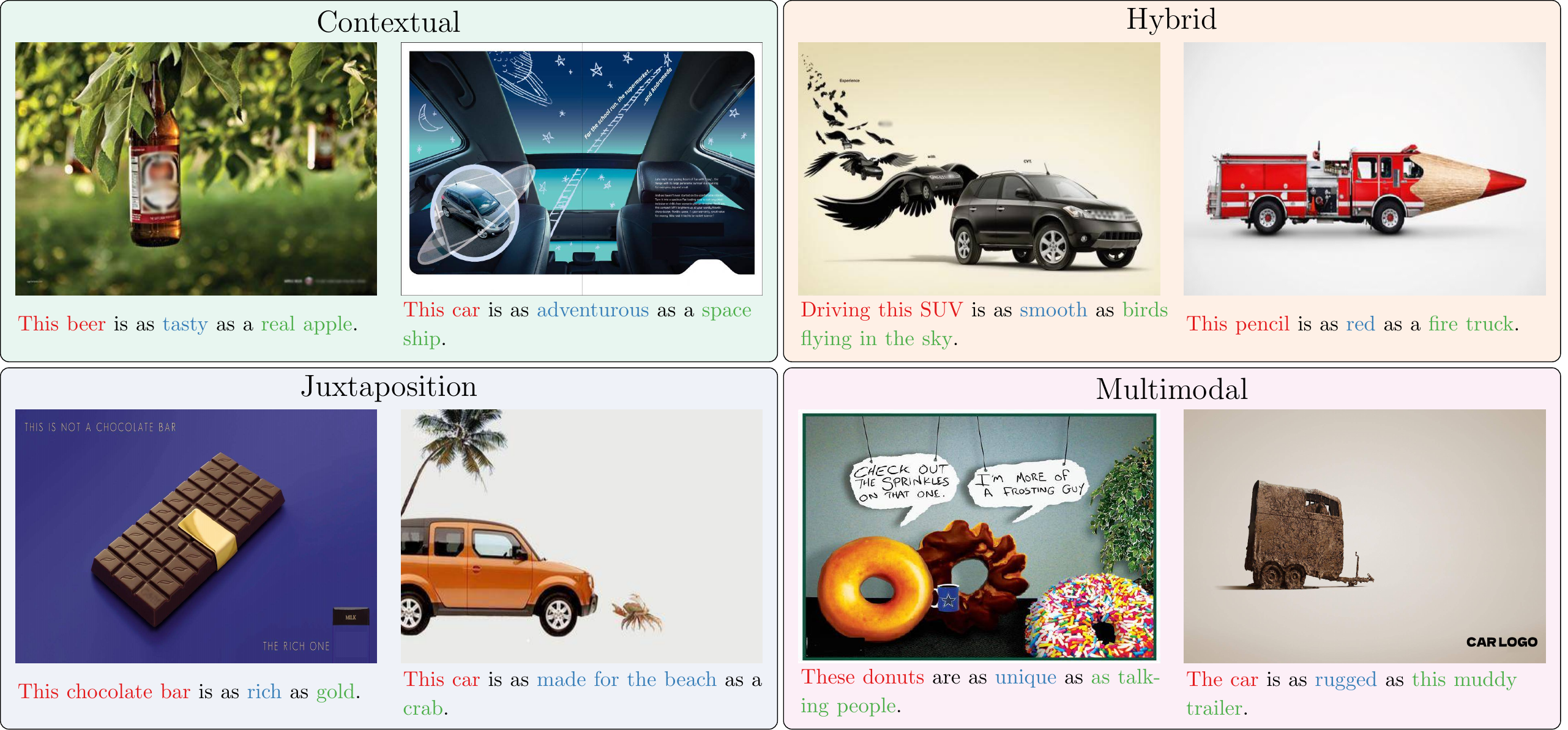} \\
\caption{\textbf{Sample Visual Metaphors with their Annotations.} There are different types of visual metaphors. The type depends on how the primary and secondary concepts are visually depicted. Here are sample Ad images from~\cite{hussain2017automatic} where we annotated the \textcolor{fontcolor1}{primary concept}, \textcolor{fontcolor3}{secondary concept} and their \textcolor{fontcolor2}{relationship}.}
\label{fig:metaphor_samples}
\end{figure*}

\vspace{0.1mm}
\noindent \textbf{Visual Metaphors} are images where the primary
and secondary concepts are visually depicted in an image conveying the metaphorical message to the viewers. Visual metaphors are widely used in mass media communications like advertising and journalism~\cite{forceville2002pictorial,stowe2021metaphor,scott1994images}.
In this work, we work with Ad images, as metaphors tend to be prevalent in ads.
There are numerous ways a metaphor can be represented visually. Following the classification in~\cite{forceville2002pictorial}, there are at least 4 different types of visual metaphors.
Fig.~\ref{fig:metaphor_samples} shows sample images that belong to these types along with our annotations of primary, secondary concepts and their relationship. In \textit{contextual} metaphors, either the primary or secondary concept is not explicitly visible, but is inferred from the context (e.g., apple in the left-most image). In \textit{hybrid} metaphors, the primary and secondary concepts are visually conflated.
\textit{Juxtaposition} forms one of the simplest visual metaphor types, where the two concepts are just presented next to each other. \textit{Multimodal} metaphors represent one of the concepts with another
modality, such as text or logo. In practice, visual metaphors use multiple of these strategies to convey a metaphor in an effective manner. In many cases, the implied metaphorical meaning is somewhat open-ended. Interpretation of visual metaphors depends on several external factors, such as familiarity with the brands and cultural context.

These visual variations and nuances make automatic cognition or generation of visual metaphors highly challenging.
While the last decade has seen rapid progress in many areas of understanding and generation tasks, prior works in computer vision focus heavily on literal interpretation of images and overlook the importance of metaphorical reasoning in understanding the image message~\cite{tong2021recent,achlioptas2021artemis}.
We believe that developing AI systems with metaphorical comprehension and generation capabilities can greatly assist humans in creative endeavors involving conveying concepts in new and exciting ways. Such systems provide an important step towards conferring human-like creativity to AI models.

To this end, we introduce multiple interesting tasks and construct metaphor annotations that enable comprehensive research on visual metaphors. As metaphors are more common in visual Ads, we start with the Pitt's Ads dataset images~\cite{hussain2017automatic} and then perform a rigorous multi-stage annotation process with expert annotators to filter metaphorical images, add metaphor annotations, and perform additional validation steps to clean the annotations\footnote{The Ads, while useful for the purposes of our paper, some images may perpetuate harmful stereotypes according to characteristics such as gender.}.
While there is recent work making advances in understanding non-literal interpretations in natural language research~\cite{choi2021melbert,chakrabarty2021mermaid}, this work proposes the first step towards metaphor analysis in images.

As illustrated in Fig.~\ref{fig:teaser}, we perform comprehensive evaluations with state-of-the-art techniques on four sets of tasks, which we call \textbf{MetaCLUE}:
1. \underline{C}lassification: This is binary classification task of estimating whether a
given image contains a metaphor or not. In other words, \textit{Are visual features indicative of whether there exists a metaphor in a given image or not?}.
2. \underline{L}ocalization: Here, the task is to \textit{localize the image regions that invoke the primary and secondary concepts in the viewer}. This is similar to a standard object detection task, but is more complicated in the case of visual metaphors as the primary/secondary concepts may not be explicitly present in an image.
3. \underline{U}nderstanding: \textit{Can our models understand the metaphorical message in a given image?} We pose this understanding problem as 3 tasks where we can quantitatively measure the performance: Retrieval, Captioning and Visual question answering.
4. g\underline{E}neration: \textit{Can we generate an image that conveys the metaphor, given the metaphorical message as a text prompt?}

We comprehensively evaluate existing state-of-the-art techniques for each of these tasks on our collected metaphor annotations. We evaluate the models both in a zero-shot manner as well as with finetuning on our annotations. Even though finetuning resulted in some improvements, most models
struggle to produce satisfactory results in many cases, demonstrating the difficulty of these tasks. Our experiments highlight several strengths and weaknesses of the existing techniques on comprehending and generating visual metaphors, providing a concrete first step towards further AI research on this fascinating topic.

\section{Related Work}
\label{sec:relatedwork}

\noindent
\textbf{Creativity and Metaphorical Abstraction.}
Creativity often involves an innovative fusion of objects, attributes, or relationships from previous knowledge to generate new concepts~\cite{bonnardel2005towards,wilkenfeld2001similarity}. Metaphors can serve as an invaluable tool for expressing creative insights and also to stimulate new ones~\cite{fauconnier2008way,indurkhya2010role}. The cognitive research community has made initial attempts in understanding different realizations of metaphors such as language metaphors and visual metaphors 
\cite{forceville2002pictorial,martin2006corpus,fussell1998figurative}. Studies of visual persuasion show that visual metaphors may be more effective than language metaphors in terms of producing a greater degree of cognitive creativity\cite{mcquarrie1999visual}. In addition to improving creativity, metaphors are also known to elicit pleasure since the initial ambiguity in (re-)conceptualizing a target entity in terms of a source stimulus generates interest and motivation, and the subsequent resolution is rewarding -- explaining the importance of using creative metaphorical processes in art, advertising, and marketing\cite{boozer1990using,lundmark2005metaphor,phillips2004beyond}.

\noindent
\textbf{Metaphors in Language Research.}
 Computational Linguistic studies show that metaphors are ubiquitous in language, occurring once per three sentences on average~\cite{steen2010method,turney2011literal,rai2020survey}. Recently, an increasing number of research efforts have explored the limitations and challenges in detecting and decoding the meaning of language metaphors\cite{chakrabarty2021mermaid,tong2021recent}. 
 While there exist several computational models~\cite{choi2021melbert,chen2020go,gong2020illinimet,terai2010computational,mason2004cormet,ovchinnikova2014generating,stowe2021metaphor,yu2019avoid} for metaphor identification, interpretation and generation in language, there exists very little work on computational modeling of visual metaphors.

\noindent
\textbf{Metaphors in Computer Vision.}
Much of computer vision literature is focused on understanding and generating literal images. Automatic metaphorical interpretations of images is highly challenging and requires multi-faceted cognitive reasoning that involves visual reasoning coupled with the use of external knowledge. Recent work on affective captioning using the ArtEmis data set \cite{achlioptas2021artemis} includes some captions that evoke metaphors to explain emotions, but that dataset is focused on visual art and does not specifically consider metaphors. 
There is no explicit prior work for comprehensive evaluation and development of models that can automatically comprehend or generate visual metaphors. 
At the same time, there exist several studies that demonstrate the potential of visual metaphors. For instance,
 
some studies\cite{barthes1977image,phillips2000impact} suggest that advertisements containing metaphorical images will be more persuasive compared to ads with language metaphors, non-metaphorical ads, or literal images.
There have been some prior computational models~\cite{indurkhya2013empirical,chilton2019visiblends} for metaphor generation, but they are not thoroughly validated against any benchmark datasets or user studies.
Some works~\cite{hussain2017automatic,ye2018advise} propose datasets and techniques for general Ad image understanding with a focus on the challenging aspects of non-literal interpretations in Ad images.
. 
However they do not explicitly collect any metaphorical annotations, nor do they provide the corresponding analysis. In this work, we start with an existing Ad image dataset~\cite{hussain2017automatic} and perform extensive human studies to filter metaphorical images and collect detailed annotations accounting various aspects of metaphoric interpretation. Prior works such as Multi-MET~\cite{zhang2021multimet} and MET-Meme~\cite{xu2022met}, propose metaphorical annotations but does not annotate the relationship between primary and secondary. In MetaCLUE, in addition to providing relationship annotations, we also collect detailed bounding box annotations that help localize the image regions invoking the primary, secondary concepts in the viewer. We further provide VQA style question and answers. 
\section{MetaCLUE}
\label{sec:dataset}

We introduce four different high-level tasks in MetaCLUE that enable comprehensive evaluation and development of visual metaphor research: \underline{C}lassification, \underline{L}ocalization, \underline{U}nderstanding and g\underline{E}neration. In the rest of this section, we first describe our annotation collection process for the tasks and next provide the analysis of using existing state-of-the-art techniques for each of these tasks.

\begin{figure}
\centering
\includegraphics[width=0.75\linewidth,height=0.53\linewidth]{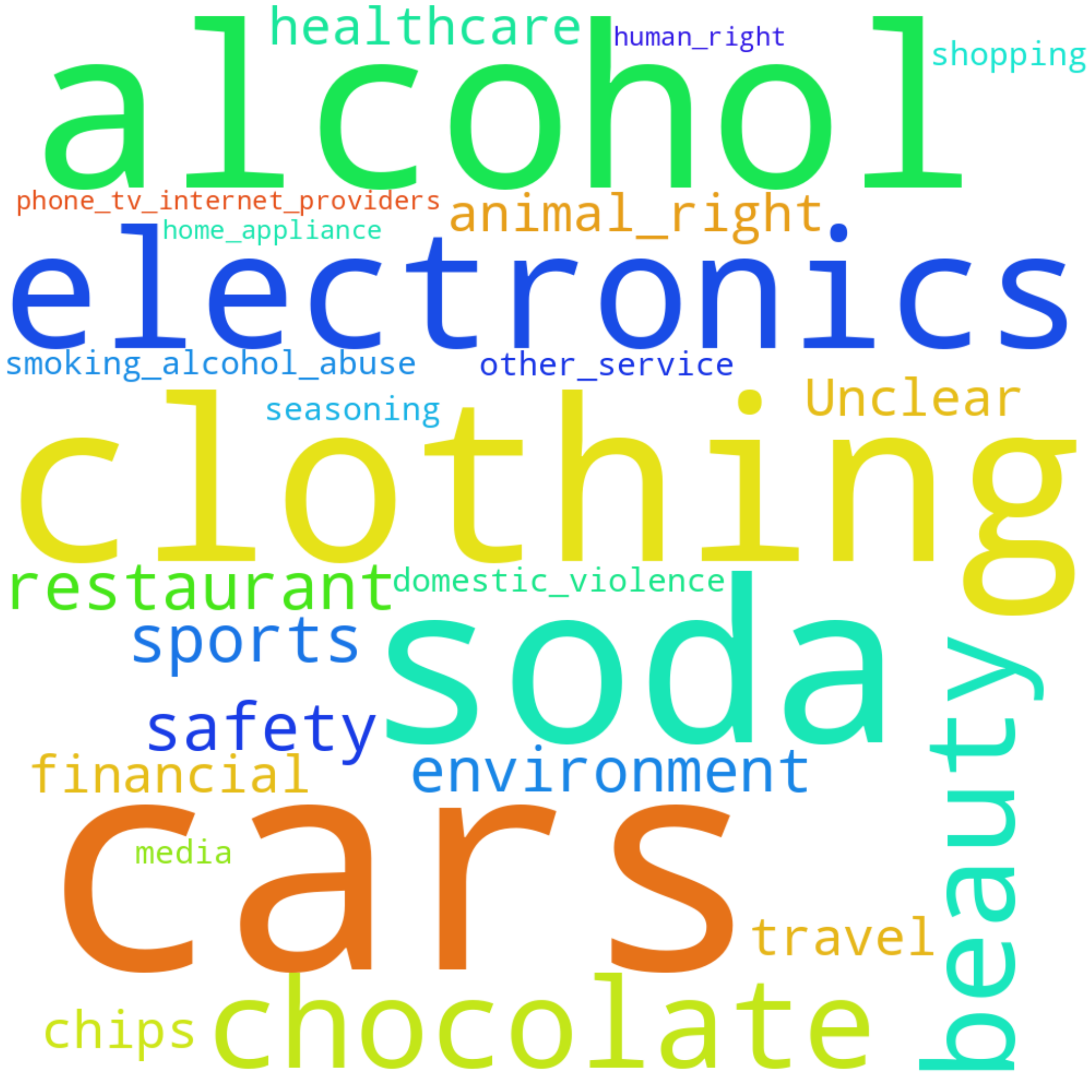} \\
\caption{\textbf{Distribution of Topics} in our annotated metaphorical images from Pitt's Ad dataset.}
\label{fig:distrib}
\end{figure}

\subsection{Metaphor Classification}
Following the fundamental vision task of image classification, we first ask: \textit{Can we develop models that can classify whether or not a given image contains a visual metaphor?} In other words, is it possible to just use visual cues to estimate whether or not there exists a metaphorical interpretation of an image?

\vspace{0.1mm}
\noindent \textbf{Annotations.}
For this task, we need to label whether or not each image contains a metaphor. Since we tend to see more metaphorical images in Ads, we start with images from an existing Ads
dataset published by the University of Pittsburgh~\cite{hussain2017automatic} and manually annotate whether a given Ad image contains a metaphor or not. Pitt's Ads dataset contains images of both product ads (e.g. phone ads) as well as public service announcements (e.g. forest conservation ads).
Concretely, to make annotations more efficient, we use a subset of 8.5K Ad images from this dataset that are annotated to have `symbolic' (fun, adventurous, etc.) references. We find this subset to contain a considerable portion of metaphors.
Specifically, we collect 5 Yes/No annotations for each of these 8K images from different annotators. To obtain
high-quality annotations, we explained what a visual metaphor is to the annotators and also conducted qualifying exams to pick final annotators for this task.
We consider the images with 3 or more `Yes' annotations as visual metaphors and the remaining as non-metaphorical.
At the end, we identify 5061 images containing visual metaphors out of 8480 images. Pitt's Ads dataset~\cite{hussain2017automatic} also
comes with topic annotations (e.g., restaurant, car, animal rights etc.). Fig.~\ref{fig:distrib} shows the word cloud plot of different topics in these 5061 metaphorical images.
We split the metaphorical images into 3730 train and 1331 test images by maintaining the same distribution of topics in both the splits.
We have two types of negative sets (non-metaphorical images) for classification experiments. One is formed by the
remaining 3419 non-metaphorical images in the symbolic set, and another one is created separately by annotating an additional 3000 literal (and non-symbolic) images from Pitt's Ads dataset~\cite{hussain2017automatic}. We add 2000 of 3419 symbolic negative images to our train split, and add the remaining 1419 images to our test split. Similarly 2000 of 3000 literal negatives are used in training, and the remaining 1000 are used for testing.

\vspace{0.1mm}
\noindent \textbf{Evaluation and Results.}
Using our collected metaphor images, we evaluate the performance of the following state-of-the-art models in classifying an input image as metaphor: EfficientNet\cite{tan2019efficientnet} and Vision Transformer (ViT)\cite{dosovitskiy2020image}. We fine-tune these models to classify metaphor vs. symbolic-non-metaphors and metaphor vs. literal, and use 20\% of corresponding train splits for validation. Test results are reported in Table~\ref{tab:classif_results}. 
Although the performance of ViT is significantly better than EfficientNet and random baselines, there is still ample room for improvement. In particular, models find it easier to distinguish metaphor images from literal images, and struggle to identify metaphors within the symbolic image pool.

\begin{table}[t]
\begin{center}
\footnotesize
\tabcolsep 3pt
\begin{tabular}{p{2cm}c  c c c c c} 
\toprule
 & & \multicolumn{2}{c}{\textit{Symbolic Neg.}} & & \multicolumn{2}{c}{\textit{Literal Neg.}}\\ 
\cline{3-4} \cline{6-7} \\[-0.7em]
Model  & $\#$Params & Val & Test & & Val & Test \\
\midrule
Random 
 & N/A & $63.10$ & $51.60$ & & $60.66$ & $57.12$\\
EfficientNet-B0 
 & 5.3M & $60.76$ & $49.67$ & & $70.94$ & $50.30$\\
EfficientNet-B7 
 & 66M & $61.44$ & $48.54$ & & $69.84$ & $49.82$\\
ViT-B/16 
 & 86M & $\mathbf{69.31}$ & $\mathbf{66.98}$ & & $\mathbf{84.04}$ & $\mathbf{81.24}$\\
ViT-L/16 
 & 307M & $65.83$ & $60.65$ & & $81.45$ & $80.52$\\
\bottomrule
\end{tabular}
\caption{\textbf{Accuracy of Metaphor Classification} (binary classification accuracy) using state-of-the-art classification architectures of EfficientNet~\cite{tan2019efficientnet} and ViT~\cite{dosovitskiy2020image}.}
\label{tab:classif_results}
\end{center}
\end{table}

\subsection{Metaphor Understanding}
\label{sec:understanding}
We now describe how we collect annotations that help in capturing the metaphorical message from the images collected in previous section. 

  \begin{table*}[t]
\begin{center}
\footnotesize
\tabcolsep 4.4pt
\begin{tabular}{lc cc c cc c cc c cc c cc c c} 
\toprule
 & & \multicolumn{2}{c}{\textit{Random Neg}} & & \multicolumn{2}{c}{\textit{Neg Prim}} & & \multicolumn{2}{c}{\textit{Neg Sec}} & & \multicolumn{2}{c}{\textit{Neg Prim+Rel}} & & \multicolumn{2}{c}{\textit{Neg Sec+Rel}} & & \textit{Swap}\\ 
 \cline{3-4} \cline{6-7} \cline{9-10} \cline{12-13} \cline{15-16}  \\[-0.5em]
Model & & p@1 $\uparrow$ & rank $\downarrow$ & & p@1 $\uparrow$ & rank $\downarrow$ & & p@1 $\uparrow$ & rank $\downarrow$ & & p@1 $\uparrow$ & rank $\downarrow$ & & p@1 $\uparrow$ & rank $\downarrow$ & & accuracy $\uparrow$ \\
\midrule
CLIP (ViT-B/16) 
& & $70.97$  & $3.49$ & & $46.67$  &  $10.11$ & & $38.14$ &  $13.97$ & & $49.36$ &  $8.80$ & & $42.25$ & $11.69$ & & $40.61$ \\
CLIP (ViT-B/32) 
& & $61.78$  & $4.19$& & $38.74$  & $11.60$ & & $33.20$ & $14.84$ & & $39.79$ & $10.29$ & & $36.27$ & $12.71$ & & $41.28$ \\
CLIP (ViT-L/14) 
& & $76.66$  & $3.17$ & & $51.75$  & $9.22$ & & $39.86$ & $13.19$ & & $54.74$ & $7.70$ & & $45.99$ & $11.02$ & & $43.08$ \\
ALBEF 
& & $39.79$  & $8.57$ & & $27.00$  & $16.53$ & & $29.31$ & $15.79$ & & $26.77$ & $15.58$ & & $28.42$ & $14.59$ & & $46.67$ \\
ALBEF (MSCOCO) 
& & $44.87$  & $7.55$ & & $31.78$  & $15.13$ & & $31.41$ & $14.09$ & & $32.53$ & $14.04$ & & $33.28$ & $12.85$ & & $48.24$\\
ALBEF (Flickr30k) 
& & $47.49$  & $8.77$ & & $35.60$  & $14.56$ & & $35.22$ & $13.66$ & & $36.35$ & $13.58$ & & $36.12$ & $12.79$ & & $\mathbf{49.81}$ \\
\midrule
FT CLIP (ViT-B/16) 
& & $76.81$  & $2.25$ & & $49.81$  & $9.65$ & & $45.47$ & $10.24$ & & $53.40$ & $8.07$ & & $50.63$ & $8.18$ & & $44.65$ \\
FT CLIP (ViT-B/32) 
& & $68.06$  & $2.82$ & & $43.00$  & $10.82$ & & $39.64$ & $11.23$ & & $44.72$ & $9.39$ & & $43.45$ & $9.35$ & & $45.69$ \\
FT CLIP (ViT-L/14) 
& & $\mathbf{81.75}$  & $\mathbf{1.78}$ & & $\mathbf{57.66}$  & $\mathbf{7.48}$ & & $\mathbf{49.06}$ & $\mathbf{9.40}$ & & $\mathbf{61.25}$ & $\mathbf{5.99}$ & & $\mathbf{57.06}$ & $\mathbf{7.33}$ & & $43.75$ \\
\bottomrule
\end{tabular}
\caption{\textbf{Performance of retrieval models} on $K$ random (column 2) and hard negative candidates (columns 3-7) ($K = 50$ for columns 2-6, $K = 2$ for last column).}
\label{tab:hn-retrieval}
\end{center}
\end{table*}

\vspace{0.1mm}
\noindent \textbf{Annotations.}
We provide detailed instructions and several examples to the annotators to help them annotate primary and secondary concepts in the metaphor and also the characteristic/relationship that is transferred from secondary to the primary. We conduct multiple pilot studies to reduce the noise and to improve inter-annotator agreement. Enforcing the annotators to make sure that their annotations are linguistically readable in the following syntactic structure helped us in improving quality and consistency of the annotations: ``\texttt{\_\_\_ is as \_\_\_ as \_\_\_}", where the first blank is the primary concept, the second blank is the relationship, and the third blank is the secondary concept. Figure~\ref{fig:metaphor_samples} shows some examples of these annotations. We collect 5 metaphor annotations for each image. As interpretations of metaphorical images can be highly subjective, there can exist more than one interpretation for each image, which makes it difficult to automatically remove noisy annotations. Therefore we conduct an additional human study where we show each of the annotation to five annotators and ask them to verify the correctness along three dimensions: (a) \textit{Is the grammar correct?}; (b) \textit{Are primary and secondary concepts correct?}; and (c) \textit{Is the relationship correct?}. We remove annotations with a low number of votes out of 5 along each of the three dimensions, resulting in a total of over 26k clean annotations.
We evaluate state-of-the-art models in understanding metaphorical message from the input images using 3 tasks namely, Retrieval, Captioning and Visual question answering.

\vspace{1mm}
\noindent \textbf{Retrieval}.\label{sec:retrieval}
The goal of this task is to retrieve the correct metaphor interpretation/statement from a candidate set given an image. In our candidate set, we choose exactly one positive (correct) metaphorical statement from its ground truth messages and uniformly sample $K-1$ random negative statements from other images. Table~\ref{tab:hn-retrieval} shows the results obtained with CLIP~\cite{radford2021learning} and ALBEF~\cite{li2021align} for $K$=$50$, reporting retrieval precision@1 and rank\footnote{Rank measures the averaged ranking value of the highest-ranked ground-truth statement with 1 being the highest possible rank.}. Although CLIP ViT-L/14 shows good zero-shot performance on random negatives with more than 76\% accuracy, we observe a large drop in performance as we increase $K$ to $\{100, 500, 1000\}$. We further fine-tune CLIP models using 70\% of metaphor annotations as the train set and see gains by up to $+7$ absolute points in p@1. 
In summary, the performance of models is impressive with less than $50$ negative candidates whereas the performance drops greatly by increasing $K$. 

We hypothesize that models might simply be looking at salient objects rather than comprehending the underlying semantics of metaphor in finding the correct candidate. To test this, we mine hard negative (HN) statements and use them as our candidate set. Specifically, we construct the following five types of HNs: (a) \textbf{Neg Prim}: candidates obtained by replacing the primary concept in the metaphor statement with the primary concept from another image\footnote{We swap objects with images having different topic to make sure that the generated HNs are actually negatives.}; (b) \textbf{Neg Sec}: replacing secondary concept likewise; (c) \textbf{Neg Prim+Rel}: replacing primary and relationship; (d) \textbf{Neg Sec+Rel}: replacing secondary and relationship; (e) \textbf{Swap Prim\&Neg}: swapping primary and secondary from the same image\footnote{With swapping $K$ is always 2.}. Table~\ref{tab:hn-retrieval} shows the results. We see a significant drop of up to 30\% by using HNs as negative statements for $K$ = $50$, indicating the difficulty in comprehending and distinguishing metaphorical abstraction of concepts. We find performance of \textbf{Neg Prim} is significantly higher than \textbf{Neg Sec}, suggesting that the models tend to rely more on primary object than secondary object in identifying correct interpretation. Overall, there is ample room for improvement indicated by the steep drop in model performance with HNs.

\begin{table*}[t]
\begin{center}
\footnotesize
\tabcolsep 2.5pt
\begin{tabular}{l r rrrrr} 
\toprule
Captioning & \multicolumn{1}{c}{Acc} & \multicolumn{1}{c}{BLEU4} & \multicolumn{1}{c}{ROUGE-L} & \multicolumn{1}{c}{METEOR} & \multicolumn{1}{c}{SPICE} & \multicolumn{1}{c}{CIDEr} \\
\midrule
Whole caption & 1.1\% & 0.254 & 0.536 & 0.220 & 0.186 & 1.076 \\
Primary & 29.9\% & 0.327 & 0.407 & 0.338 & 0.244 & 0.931 \\
Secondary & 13.7\% & 0.249 & 0.307 & 0.193 & 0.155 & 0.550 \\
Relationship & 23.6\% & 0.485 & 0.203 & 0.226 & 0.028 & 0.296\\
\bottomrule
\end{tabular}
\hspace{8pt}
\begin{tabular}{l r rrrrr} 
\toprule
\multicolumn{1}{c}{VQA} & \multicolumn{1}{c}{Acc} & \multicolumn{1}{c}{BLEU4} & \multicolumn{1}{c}{ROUGE-L} & \multicolumn{1}{c}{METEOR} & \multicolumn{1}{c}{SPICE} & \multicolumn{1}{c}{CIDEr} \\
\midrule
All questions & 19.9\% & 0.329 & 0.286 & 0.249 &  0.185 & 0.851 \\
Primary & 21.5\% & 0.291 & 0.348 & 0.277 & 0.290 & 1.099 \\
Secondary & 12.8\% & 0.238 & 0.232 & 0.181 & 0.234 & 0.735 \\
Relationship & 25.6\% & 0.449 &0.275 & 0.285 & 0.038 & 0.706\\
\bottomrule
\end{tabular}
\caption{\textbf{Metaphorical image captioning (left) and visual question answering (right) performance of PaLI}~\cite{pali}. We report different metrics (higher the better) using both the exact match accuracy (Acc) and standard text generation metrics. For image captioning, we evaluate the whole predicted caption as well as parsed primary object, secondary object, and relationship. For VQA, we evaluate on all predicted answers as well as provide the breakdown for each question type.}
\label{tab:capvqa-results}
\end{center}
\end{table*}

\begin{figure}
\centering
\includegraphics[width=\linewidth]{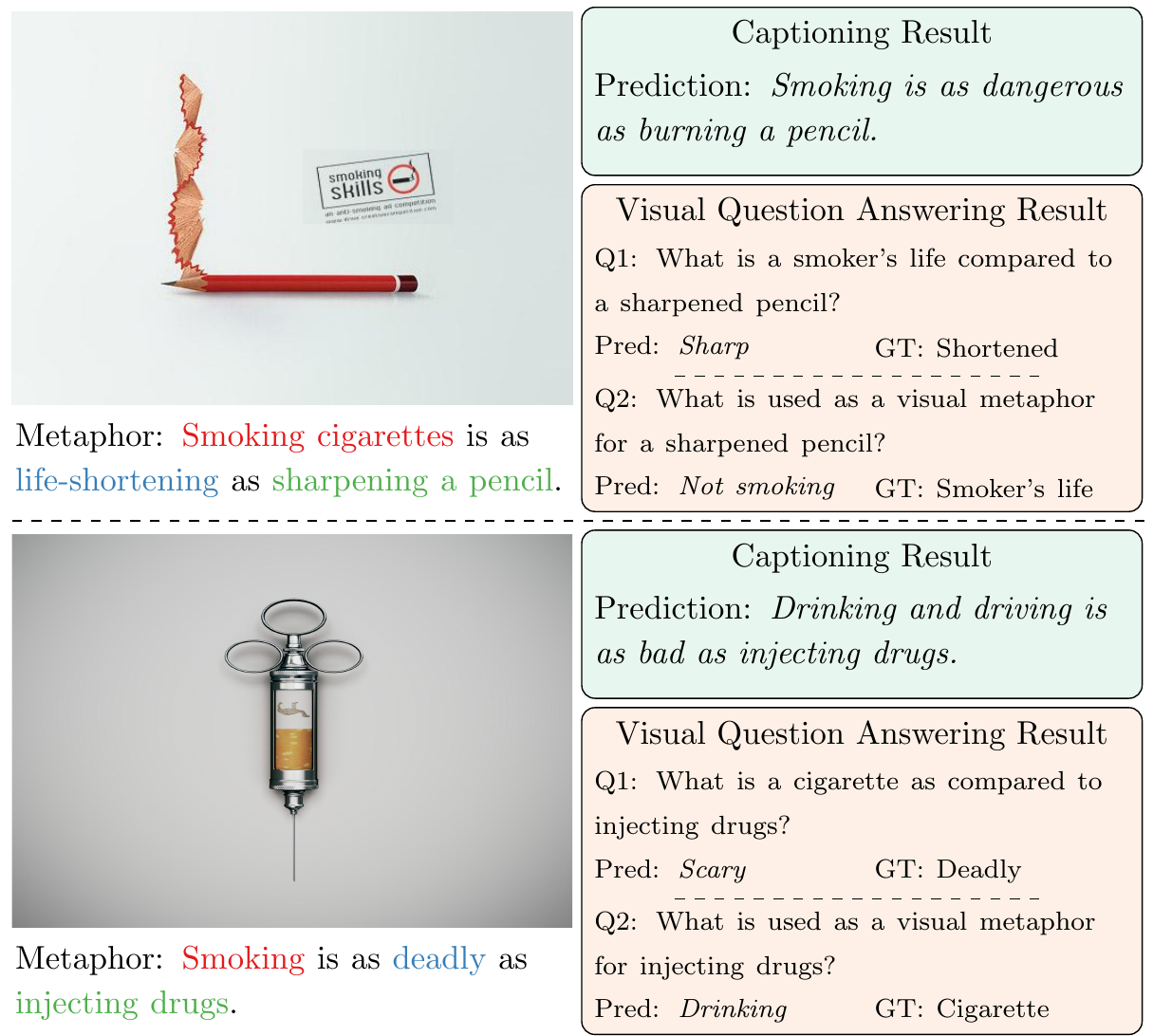}
\caption{\textbf{Results of PaLI~\cite{pali}} for captioning and visual question answering on sample images in our test split.}
\label{fig:understanding_qual}
\end{figure}

\vspace{1mm}
\noindent \textbf{Captioning}.

Here, we propose metaphor image captioning task, where the input is an image and the target is the metaphorical message in the syntactic structure \texttt{$<$primary\_concept$>$ is as $<$relationship$>$ as $<$secondary\_concept$>$}. We fine-tune and evaluate the state-of-the-art literal image caption model PaLI-17B~\cite{pali} based on the exact match accuracy (maximum over all references) and standard metrics for image captioning BLEU4~\cite{bleu}, ROUGE-L~\cite{rouge}, METEOR~\cite{meteor}, SPICE~\cite{spice}, and CIDEr~\cite{cider}. 
Since the target captions follow the fixed syntactic structure, we parse each predicted caption into the primary concept, the secondary concept, and their relationship, and use the same set of metrics for the whole caption to evaluate.\footnote{The score of 0 is given if parsing fails.}
Table~\ref{tab:capvqa-results} (left) summarizes the results and Figure~\ref{fig:understanding_qual} provides sample qualitative results.
We observe that PaLI generally struggles on this task. For instance, it achieves a CIDEr score of 1.076, compared to 1.491 for state-of-the-art literal image captioning on the popular COCO-Captions~\cite{cococap}. Further, the model struggles with predicting the target relationship when the metrics favor recall (e.g., ROUGE-L) and with predicting the target secondary concept when the metrics favor precision (e.g., BLEU).

\vspace{1mm}
\noindent \textbf{Visual Question-Answering (VQA)}.
We propose metaphorical open-ended (i.e., not vocab-based) VQA task, where the input is an image and a given question, and the target is the answer. We use fixed templates to generate 2 VQA questions whose answer is the primary concept, 2 for secondary concept, and 2 for relationship. Again, we fine-tune and evaluate the state-of-the-art literal VQA model PaLI-17B~\cite{pali}, using the same set of metrics as in image captioning.
Table~\ref{tab:capvqa-results} (right) summarizes the results and Figure~\ref{fig:understanding_qual} provides sample qualitative results.
Overall, we find that PaLI performs poorly, only achieving the average accuracy score of 19.9\%, while the state-of-the-art literal VQA on the popular VQAv2~\cite{vqa2} benchmark is 77.6\% on ``other'' questions. Additionally, the model struggles the most with answering questions that ask for the secondary concept.

\begin{figure*}[t!]
\centering
\includegraphics[width=\linewidth]{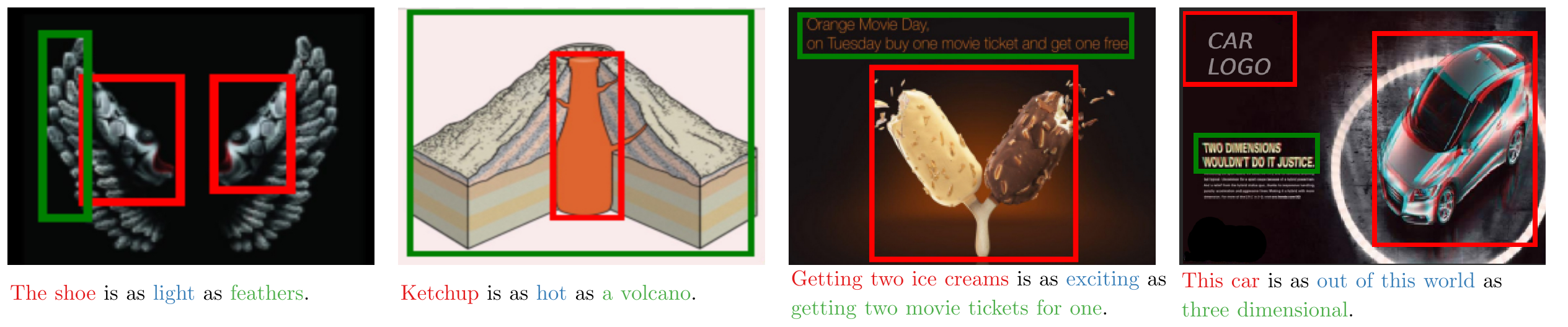} \\
\caption{\textbf{Sample Localization annotations} showing annotated bounding boxes around primary and secondary concepts. Notice the diversity of types in bounding boxes: explicit, contextual, logo and texts. This makes metaphor localization more challenging compared to standard object detection.}
\label{fig:localization-examples}
\end{figure*}

\begin{figure}[t!]
\centering
\includegraphics[width=0.98\linewidth]{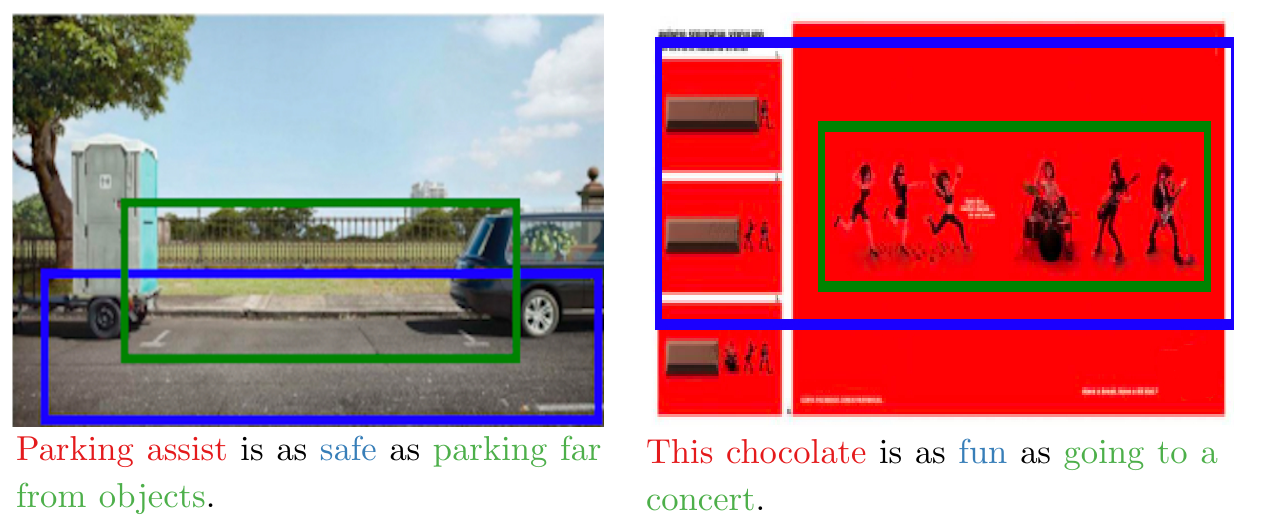} \\
\caption{\textbf{Sample Localizations} with the phrase-grounding model from~\cite{li2022adapting}, where the secondary concepts are contextual. GT boxes are shown in green, whereas the predictions are shown in blue.}
\label{fig:localization-results}
\end{figure}

\subsection{Metaphor Localization}
\label{sec:localization}
Here, the task is to localize the image regions that invoke either the primary or secondary concept in the viewer.
This is similar to the phrase grounding task of localizing objects using
free-form natural language phrases~\cite{kazemzadeh2014referitgame,mao2016generation,yu2018mattnet}, but with some key differences due to the
peculiarities of visual metaphors in comparison to literal images used in standard vision datasets.

\vspace{0.1mm}
\noindent \textbf{Annotations.}
As discussed earlier, there are diverse types of metaphors based on how the primary and secondary concepts are visually depicted in an image (see Fig.~\ref{fig:metaphor_samples}).
There are at least two key differences in metaphor localization compared to standard localization in literal images: 1. A given concept can be present in the image either explicitly or in a contextual manner (for e.g., contextual visual metaphors in Fig.~\ref{fig:metaphor_samples}). 2. Visual metaphor Ads are inherently multimodal and a concept can be invoked by other modalities such as text or logo. See the multimodal visual metaphors in Fig.~\ref{fig:metaphor_samples}.
As a result, we not only annotate the bounding boxes that invoke the primary/secondary concept in the viewer, but we also annotate the type of that bounding box. A bounding box can be of one of the 4 types: Explicitly present, Contextually present, Logo or a Text.
Specifically, for each of the 5061 metaphorical images, we pick the best metaphor annotation (primary, secondary concepts and their relationship) according to their validation scores (see previous section) and
collect bounding box annotations for both the primary and secondary concepts.
We collect all the bounding boxes that invoke both the primary and secondary concepts and also their type (explicit, contextual, logo or text) for each of the images.
We use 5 annotators for each annotation and choose the bounding boxes with the best inter-annotator agreement. We use the same train and test splits as used in understanding tasks. Fig.~\ref{fig:localization-examples} shows sample localization annotations. 
We collected over 30k bounding box annotations for this task on 5061 metaphor images.

\vspace{0.2mm}
\noindent \textbf{Detection}.
Recent detection and localization models~\cite{li2022adapting,zhou2021denseclip} pre-trained on image and caption pairs are shown to achieve remarkable localization performance on discriminating fine-grained objects and unseen concepts. Specifically, we evaluate \cite{li2022adapting} which leverages the effective
image representations in CLIP by extracting spatial features from it. Using these spatial features, for each pixel location, the model computes the inner product between
the spatial feature and the phrase embedding extracted from CLIP to predict the bounding box. In our case, we pass the primary or secondary concept as input phrase to~\cite{li2022adapting} the estimate the corresponding bounding box. Table~\ref{tab:localization-results} summarizes the detection results (using Mean Average Precision) on our test split. We find relatively better performance in localizing secondary objects compared to primary objects. 

We show few qualitative results in Figure~\ref{fig:localization-results}.
It is worth noting that our collected annotations allow for more comprehensive analysis on localization tasks due to the availability of different types of bounding boxes (explicit, contextual, logos and texts).

\begin{table}[t]
\begin{center}
\small
\tabcolsep 4.5pt
\begin{tabular}{lcc} 
\toprule
\textit{} & mAP$_{50}$ & mAP$_{70}$ \\ 
\midrule
Primary concept & $33.22$  & $14.25$ \\
Secondary concept & $43.54$  & $31.23$ \\
\bottomrule
\end{tabular}
\caption{\textbf{Localization results} with CLIP based phrase localization model~\cite{li2022adapting} on our test split.}
\label{tab:localization-results}
\vspace{-4mm}
\end{center}
\vspace{-3mm}
\end{table}

\subsection{Metaphor Generation}
\label{sec:generation}

Recent large-scale text-to-image (T2I) generative models show remarkable success in generating highly realistic images from text prompts. \textit{Can these models also work well in metaphorical image generation?}
We evaluate two state-of-the-art generative models (Imagen~\cite{saharia2022photorealistic}, Stable diffusion~\cite{rombach2022high} (SD)) using 300 samples from MetaCLUE test set where we use the text prompts: ``An advertisement where \textit{primary-concept} is as \textit{relationship} as \textit{secondary concept}."
In addition, we finetune the stable diffusion model on our train split (same split as in Sec.~\ref{sec:understanding}).

\begin{figure}
\centering
\vspace{-2mm}
\includegraphics[width=\linewidth]{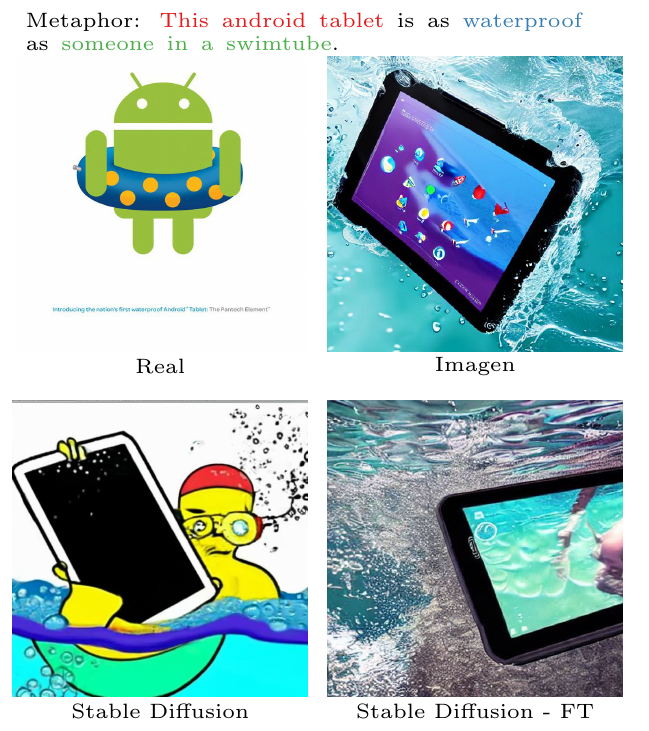}
\caption{\textbf{Sample Image Generations} for a given metaphorical message (shown on top) with Imagen~\cite{saharia2022photorealistic}, Stable Diffusion~\cite{rombach2022high} and fine-tuned (FT) version of Stable Diffusion.}
\label{fig:generation-examples}
\end{figure}

\vspace{0.1mm}
\noindent \textbf{Results.}
Fig.~\ref{fig:generation-examples} shows sample visual results from different T2I models, with the metaphor annotation shown on the top. The generated images capture different aspects of the metaphor (tablet, waterproof), but not the entire metaphorical message.
We compute standard metrics to automatically evaluate the quality of the generations.
Tab.~\ref{tab:generation-results} shows the standard FID~\cite{heusel2017gans} and CLIP-Similarity~\cite{radford2021learning} scores of different models. FID score evaluates the image distribution similarity between the
generation images with the corresponding real image distribution. FID scores in Tab.~\ref{tab:generation-results} shows that Imagen performs slightly better than SD. And, there is a slight improvement in FID with finetuning (SD-FT vs. SD).
In general, high FID scores in Tab.~\ref{tab:generation-results} indicate the large distribution gap between the generated and real images. CLIP similarity score, on the other hand, measures the prompt fidelity - similarity between the generated image and the corresponding input text prompt according to the CLIP model~\cite{radford2021learning}.
CLIP similarity scores in Tab.~\ref{tab:generation-results} follow the similar trend as FID scores: Imagen performs better than SD in terms of the prompt fidelity, and finetuning SD model improved its prompt fidelity.

\begin{figure}[t!]
\centering
\includegraphics[width=\columnwidth]{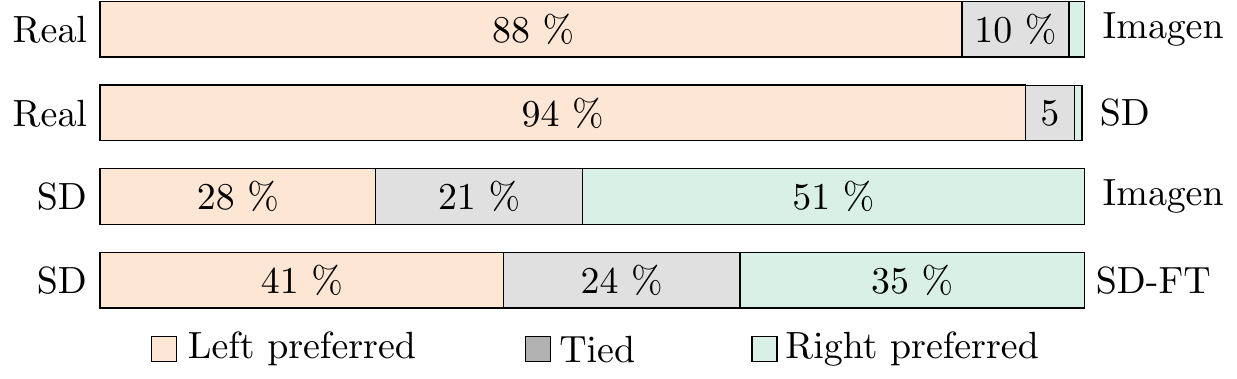}
\caption{\textbf{User Study on Image Generation Results.} Percentage of results users preferred across real, Imagen~\cite{saharia2022photorealistic}, Stable Diffusion (SD)~\cite{rombach2022high} and its fine-tuned version (SD-FT) results. Users are asked to choose the image that better depicts a given metaphor.}
\vspace{-2mm}
\label{fig:user_study}
\end{figure}

Given that both FID and CLIP scores are not tailored towards metaphorical images, these metrics are not reliable in assessing whether the generated images capture the essence of visual metaphors.
To analyze this, we perform human studies comparing two different models at a time. 
Specifically, we show a metaphorical message and the corresponding generations from two models and ask the users to pick an image that best conveys that metaphorical message.
We obtain 7 user ratings for each image pair and consider 3-out-of-7 or 4-out-of-7 to be ties.
Fig.~\ref{fig:user_study} shows the percentage of user preferences across different pairs of results. User studies also indicate that Imagen performed better than SD. An interesting finding is that finetuning SD resulted in slightly worse user preference compared to base SD model (SD-FT vs. SD). This is in contrast to standard FID and CLIP metrics that improved with finetuning.
We hypothesize that finetuning SD resulted in more realistic Ad images, but the resulting model may have forgotten some of the metaphorical priors. This calls for more effective finetuning strategies with the limited training datasets, which forms an important future work.
In addition, both Imagen and SD performed quite poorly compared to real images in conveying metaphorical messages. Real images are preferred around $88\%$ of time over Imagen results. This illustrates the big scope of improvements in generating visual metaphors.

\begin{table}[t]
\begin{center}
\footnotesize
\tabcolsep 4.5pt
\begin{tabular}{lcc} 
\toprule
\textit{Model} & \textit{FID}~$\downarrow$ & \textit{CLIP Similarity}~$\uparrow$ \\ 
\midrule
Imagen~\cite{saharia2022photorealistic} & $153.1$  & $32.1$ \\
Stable Diffusion~\cite{rombach2022high} & $161.6$  & $30.8$ \\
Stable Diffusion - FT & $154.3$  & $32.0$ \\
\bottomrule
\end{tabular}
\caption{\textbf{Analysis of Image Generation} results with standard metrics of FID~\cite{saharia2022photorealistic} and CLIP similarity~\cite{radford2021learning} scores.}
\label{tab:generation-results}
\vspace{-4mm}
\end{center}
\vspace{-5mm}
\end{table}

\section{Conclusion}
\label{sec:conclusion}
In this paper, we present a step towards comprehensive evaluation of progress on visual metaphor research. Specifically, we propose a collection of tasks related to comprehending and generating visual metaphors using AI techniques. 
Our MetaCLUE tasks include Classification, Understanding (Retrieval, Captioning, VQA), Localization and Generation.
For comprehensive evaluations, we collected high quality and rich annotations that facilitate the measurable progress.
Existing methods demonstrate poor results in many cases with our experimental analysis shedding light on strengths and drawbacks of different approaches paving a path for future research in this fascinating field.


{\small
\bibliographystyle{ieee_fullname}
\bibliography{egbib}
}

\end{document}